# Fast Video Retargeting Based on Seam Carving with Parental Labeling

Chuning Zhu

*Author Background: Chuning Zhu grew up in China and currently attends The Experimental High School Attached to BNU, in Beijing, China. His Pioneer seminar program was in the field of computer science and titled "Computers That See: Exploring New Techniques of Computer Vision."*

**Abstract**

Seam carving is a state-of-the-art content-aware image resizing technique that effectively preserves the salient areas of an image. However, when applied to video retargeting, not only is it time intensive, but it also creates highly visible frame-wise discontinuities. In this paper, we propose a novel video retargeting method based on seam carving. First, for a single frame, we locate and remove several seams instead of one seam at once. Second, we use a dynamic spatiotemporal buffer of energy maps and a standard deviation operator to carve out the same seams in a temporal cube of frames with low variation in energy. Last but not least, an improved energy function that considers motions detected through difference method is employed. During testing, these enhancements result in a 93 percent reduction in processing time and a higher frame-wise consistency, thus showing superior performance compared to existing video retargeting methods.

## 1 Introduction

As more and more smart devices are becoming a part of our lives, multimedia content, especially videos, is viewed on a variety of displays with aspect ratios that run the gamut from 1:1 to 21:9. However, videos are usually generated in a specific resolution with a set aspect ratio. Presenting them on different displays without optimization will result in a compromised visual perception. Therefore, video adaptation algorithms are of great importance to present-day multimedia consumption.

Seam carving, or dynamic resizing, is a content-aware image resizing algorithm first proposed by Shai Avidan and Ariel Shamir in 2007 (Avidan and Shamir, 2007). Unlike standard image scaling, which corrupts the ratio of content, or image cropping, which only discards peripheral pixels, seam carving maintains the salient regions of an image by recursively removing or inserting the seam with the lowest cumulative energy. A seam is an 8-connected path that crosses the image either vertically or horizontally. The cumulative energy of a seam is defined via an energy map, or a visual saliency map. The map assigns a value to each pixel according to an energy function (gradient energy, entropy etc.) and the cumulative energy of a seam is the sum of all pixel energy values in its path. Under optimal conditions, seam removal retains the salient regions of an image while creating barely noticeable artifacts.

Despite its satisfactory performance on a single image, the original seam carving algorithm is not ideal for video retargeting. First, the fact that it is based on dynamic programming renders the algorithm excessively time-consuming for the purpose of video retargeting. Resizing an individual image can be done in a reasonable amount of time; video retargeting, however, requires the adaptation of so vast a number of frames that the length of a video and the time needed for its retargeting is extremely disproportional. Second, the result of applying seam carving to separate frames is visually inconsistent. Since the seam calculation of each frame is independent of others, there are conspicuous fluctuations in the output which greatly sabotage the visual appeal. For instance, an edge may move left or right in different frames since the locations of seam removal may change.

In this paper, we present a seam-carving-based video retargeting algorithm that runs in reasonable time while retaining frame-wise consistency. We propose a multiple seam searching and removing method based on what we call Parental Labeling (SCPL). It can be demonstrated that all the seams are derivations of some parents in the first row or column. By labeling all the children (pixels in the last row or column) with values corresponding to their parents and subsequently removing the minimal child of each parent, multiple seams can be removed in one iteration. Based on the SCPL, we establish connections between frames by creating a dynamic spatiotemporal buffer, whose size is determined by dynamically calculating the average standard deviation of energy map. Once the STD is above a dynamic threshold, the buffer is fixed and the same seams derived from a uniform energy function are removed across the buffer. Finally, an improved visual saliency map that takes motion into account is employed to stabilize and smoothen the motion.

The rest of the paper is organized as follows. We first review related work concerning improved video retargeting. Then, we elucidate the proposed method in detail. Finally, we evaluate the results and conclude the paper with an analysis of pros and cons and a discussion of future work.

## 2  Related Work

Previous research has already proposed manifold video retargeting approaches based on seam carving. M. Rubinstein et al. (2008) first refined the original seam carving process to extend its functionality to video retargeting. To achieve this, they introduced a new seam carving method based on graph cut. In this method, a seam is a connected and monotonic S/T cut of the graph representation of an image. Formally, a vertical cut is a continuous function over the relevant domain $S: row \times time \rightarrow column$, and a cut in video is a 2D manifold of a 3D time-frame volume. Another enhancement is the employment of forward energy. The original backward energy calculation, which removes seams with the least energy, may increase the total energy of the image as new neighbors are created. Taking this factor into account, they proposed an algorithm that removes the seam whose removal will insert the minimal amount of energy. There are, however, limitations on this algorithm. Although it protects the structure of the image quite well, it sabotages the shape of content. Furthermore, graph cut has a higher time complexity than dynamic programming, which deviates from what we are exploring in this paper.

As for accelerated video retargeting, Stephan Kopf et al. (2009) proposed a novel seam-carving-based algorithm to adapt videos to mobile screens named Fast Seam Carving for Size Adaptation of Videos (FSCAV). FSCAV combines camera compensation with background aggregation. It also features a criterion for determining robust seams (i.e., seams in the background that can be removed from frames temporally). Another mechanism proposed by Kopf is the measuring operator of video adaptation. If the resizing factor goes beyond a threshold, other resizing methods such as scaling and cropping are introduced. As for evaluation, the FSCAV is indeed significantly faster than graph cut methods, but it still requires more than 30 minutes for typical videos with appropriate resolution. A great portion of running time is spent on live analysis of robust seams. The approach is generally limited by camera movements, but the background aggregation technique provides inspiration for our approach.

Other approaches usually modify seam carving for higher parallelism and resort to GPU for multi-threading. Kim et al. (2014) proposed an improved seam carving that is modified to run in parallel on multiple GPUs. Their method has three major novelties. First, they propose Non-Cumulative Seam Carving, an approach that resembles a greedy algorithm, to replace the traditional dynamic programming. The algorithm calculates the optimal path for each pixel simultaneously, depending exclusively on the energy values of its neighbors in the following row or column. The major purpose of this algorithm is to parallelize the calculation of the seam map. Then, the authors proposed the Adaptive Multi-Seam Algorithm. Unlike traditional SC that discards one seam per iteration, this algorithm searches for multiple neighboring seams with distances less than a threshold, and removes several seams at a time. Lastly, they elaborate on the syncing of multiple GPUs.

Another GPU-based method is proposed by C. Chiang et al. (2009) The major contributions of this method are seam-split and JND (Just Noticeable Distortion) based saliency map. The former enables the algorithm to remove multiple seams at a time by splitting "local seams" that branch off from "global seams," while the latter improves the quality of video to make it more visually appealing. With GPU acceleration, the algorithm can achieve a frame rate ranging from 10 to 30 fps, depending on the resolution of test cases.

Both of the GPU-based methods are impractical for wide application, as the majority of mobile and immobile display devices do not possess such powerful multi-threading capability. Nevertheless, they do enlighten us about possible aspects and directions of optimization.

Recently, work concerning improved visual saliency map optimized for video retargeting has emerged. Duan-Yu Chen et al (2011) proposed a novel approach based on visual saliency cube to retarget videos. First, it detects the salient points in spatiotemporal domain using modified Harris Detector. Then, using these salient points as a seed for searching, the method constructs a motion attention map where consistent motions correspond to higher value. Finally, seam carving is imposed on the motion attention map. Since moving areas have higher values, they are protected from removal. Although this approach prevents discontinuities in motion that result from ordinary seam carving, its limitation lies in the assumption that there is no camera movement. In other words, the camera has to be fixed in order for the searching algorithm to function. Detection of motion is of great importance in video retargeting as it maintains the coherence of moving objects. In our paper, a more simplistic motion-aware visual saliency map is used for the purpose of fast video retargeting.

## 3  Fast Video Retargeting Based On SCPL
### 3.1  Seam Carving With Parental Labeling (SCPL)

According to the seam carving algorithm, a seam is defined as an 8-connected path that cuts through the image either vertically or horizontally, and contains one and only one pixel in each row or column respectively. Formally, let $I$ be an $n \times m$ image, and a vertical seam is defined via:

$$S^x = \{s_i^x\}_{i=1}^n = \{(x(i), i) | i \in \{1, \ldots, n\}, |x(i) - x(i-1)| \leq 1\}, \quad (1)$$

where $x$ is a mapping $x: [1, \ldots, n] \to [1, \ldots, m]$.

Similarly, if $y$ is a mapping $y: [1, \ldots, m] \to [1, \ldots, n]$, a horizontal seam is defined via:

$$S^y = \{s_j^y\}_{j=1}^m = \{(j, y(j)) | j \in \{1, \ldots, m\}, |y(j) - y(j-1)| \leq 1\}. \quad (2)$$

Since vertical and horizontal seam carving are largely congruent, and horizontal seam carving can be done by rotation and vertical seam carving, for the sake of simplicity, the illustrations below will only address vertical seam carving.

Given an energy function $e$, the cost of a seam is defined as $E(S) = \sum_{i=1}^n e(P(S_i))$, where $P(S_i)$ is the position of the $i$th pixel along the path of a seam and $e$ is the energy of a pixel. Typically, we want to search for and remove the seam with the minimal cost, or $S_{min}$ such that:

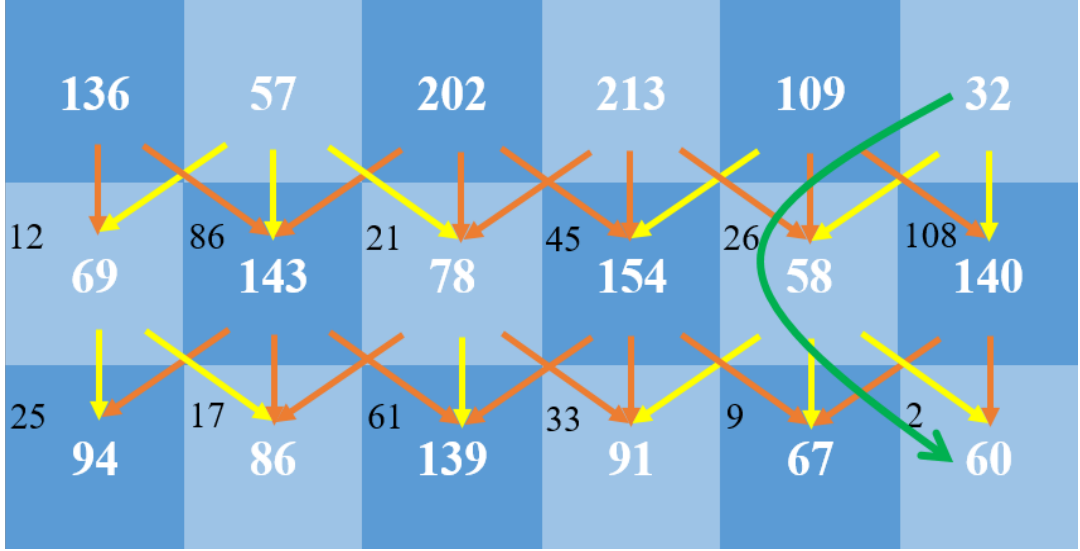

***Figure 1.*** *Illustration of the seam searching process. Black numbers are pixel energy; white numbers are cumulative energy. Orange arrows indicate eliminated paths; yellow arrows indicate selected paths; the green arrow indicates the seam with lowest energy.*

$$E(S_{min}) = \min\{E(S)\} = \min\left\{\sum_{i=1}^n e(P(S_i))\right\}. \quad (3)$$

The searching process is done through dynamic programming. We traverse the image row by row, and for each pixel $(i, j)$ in a row, a selection is made between three possible paths above it, namely $(i-1, j-1)$, $(i-1, j)$, and $(i-1, j+1)$. The pixel is connected to the path with lowest cumulative energy, ensuring that the updated cumulative energy is the least among all the possibilities. Formally, the cumulative energy is updated via the following function:

$$CE(i,j) = e(i,j) + \min(CE(i-1, j-1), CE(i-1, j), CE(i-1, j+1)). \quad (4)$$

At the end of traversal, the least cumulative energy in the last row indicates the least significant seam. During the traversal, an array of indicators is created to record the path. Seam removal can thus be done through tracing back the indicators. Since a seam is monotonic (i.e. there is exactly one pixel in each row or column), seam removal has only a local effect on the image.

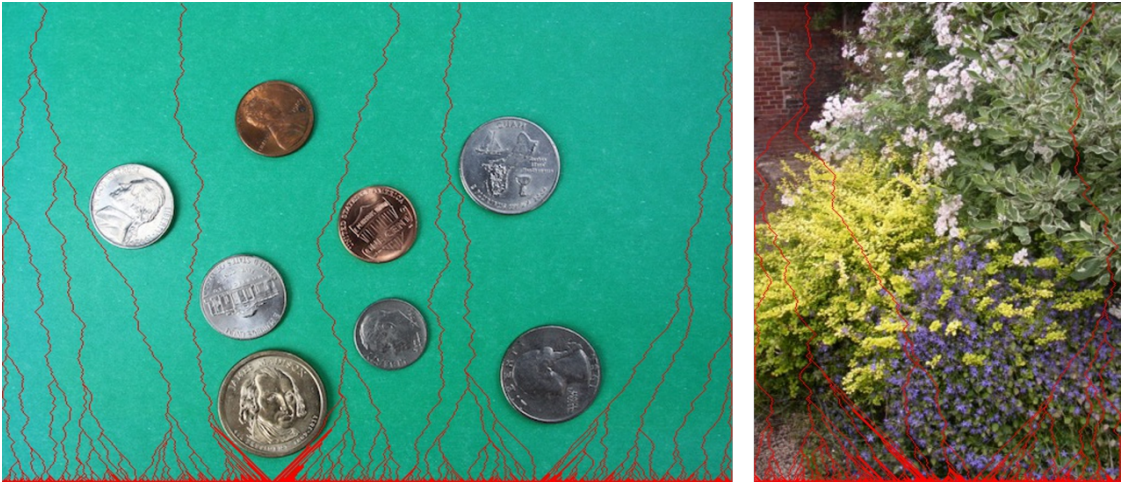

*Figure 2. Presentation of seams in images.*

Note that the collection of seams can be regarded as a mathematical function. We can define the collection of pixels in the last row as set L, and the collection of pixels in the first row as set F. Then the collection of seams can be represented as function SE: L→F. Recall that a function is a relation between sets that satisfies left-totality and right–uniqueness. Define relation SE between L and F as $\{(l, f) \in L \times F \mid l$ and $f$ are connected via a seam$\}$, the proofs of the two properties of functions are as follows:

a. Left-totality: During the last iteration of traversal, by formula (4), each pixel from the last row is connected to one in three paths from above, which is connected to some parents. $\forall l \in L, \exists f \in F \ s.t. \ lSEf$. The relation is thus left-total.
b. Right-uniqueness: Each seam connects one pixel from the last row with exactly one pixel from the first row. $\forall l \in L, \forall f, g \in F, (lSEf \wedge lSEg) \Rightarrow (f = g)$. The relation is thus right-unique.

Observe that the set containing pre-images of each element in the domain of SE is a partition of domain (pixels in the last row). In other words, each pixel in the last row can be classified by its origin into discrete categories. Specifically, let $a$ be in the domain of SE, L, a partition $P(L)$ of L is defined as:

$$P(L) = \{\text{Preim}_{SE}(\{a\}) | a \in Im_{SE}(F)\}. \tag{5}$$

Therefore, we can label each element of L (each pixel in the last row) with its image (pixels in the first row), or what we figuratively name "parent". Each child has one and only one parent, while each parent has multiple children. Since all the elements of partition $P(L)$ are pairwise disjoint, there is no overlap between the children of any two parents. After the removal of a child from one parent, the children of all other parents remain intact. In addition, since for each pixel in the image there is only one set path that can be traversed backwards, no two seams have intersection with each other. (If two seams intersect, then for the intersection point there are two possible paths to trace back to, which is impossible.) It is thus ensured that the output has uniform width. With all the restrictions clarified, it is viable to remove the child with the least cumulative energy from each parent without jeopardizing the overall structure of an image. The indices of target seams can be using through the following function:

$$Idx(I) = \{argmin(CE(a)) | a \in P(L)\}, \tag{6}$$

where $I$ is the input image and $P(L)$ is the partition defined above. With parental labeling, multiple seams can be removed at a time. The processing time is thus accelerated to various extents depending on the input.

## 3.2 Spatiotemporal Buffer Of Energy

The dynamic programming nature of seam carving casts a limit on optimization of single-frame retargeting. When we expand the scope to video retargeting, however, there exist more possibilities of improvement. Through observation, we notice that videos usually include short periods with slight variation of content. Applying seam carving separately on each frame in such a period is redundant, as the calculations are highly repetitive. Moreover, the barely noticeable differences lead to minor alterations of seam positions, which in turn accumulate to visible displacement of content between processed frames, i.e. jittering. In this paper, we propose a dynamic spatiotemporal buffer whose elements can be applied with seam carving uniformly. Figuratively, a spatiotemporal buffer is a cube whose cross sections are frames. There are two problems to be addressed about this construction: the size of the buffer and its energy map.

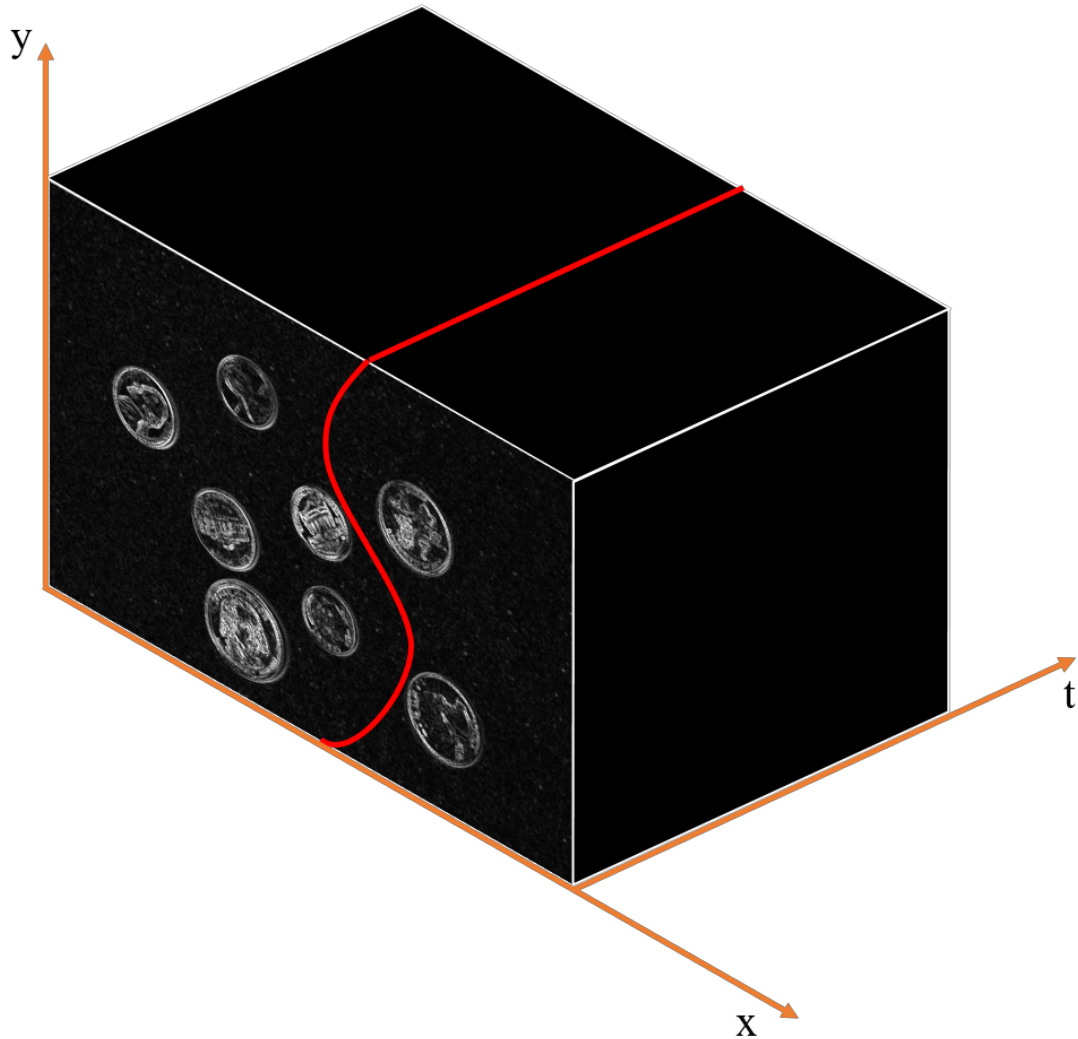

*Figure 3.* Spatiotemporal buffer of energy maps.

To determine the size of the buffer, we present an operator based on average standard deviation of energy (ASDE). Typically, frame-wise difference is directly proportional to variation of energy, i.e. lower frame-wise difference corresponds to lower variation of energy. Therefore, we construct two arrays, one as a container of color frames, the other as the spatiotemporal buffer of energy maps. The initial lengths of the two arrays are zero, yet they grow simultaneously. With each increment of frame, an ASDE is defined as follows.

$$ASDE(I) = \frac{1}{n \times m} \sum_{i=0}^{n} \sum_{j=0}^{m} Std(i,j), \quad (7)$$

where $I$ is an energy map with size $n \times m$.

For each pixel $(i,j) \in \{1,...,n\} \times \{1,...,m\}$, its standard deviation in a spatiotemporal buffer $Std(i,j)$ with length T is defined via:

$$Std(i,j) = \sqrt{\frac{1}{T} \sum_{t=0}^{T} (e((i,j)_t) - \mu_{e(i,j)}^T)^2}, \quad (8)$$

where e is the energy of a pixel and $\mu$ is the average energy of a pixel across time window $T$.

Using the ASDE as an indicator of variation, we can thus determine the size of the spatiotemporal buffer by comparing its value to a threshold. However, note that the threshold is varying, for the same upheaval has less effect on the ASDE of a larger buffer than that of a smaller one. To visualize the relation, we plot the ASDE of a spatiotemporal buffer with first $n$ consecutive frames of lowest energy and one frame of highest energy against the variable $n$, as shown in Figure 4.

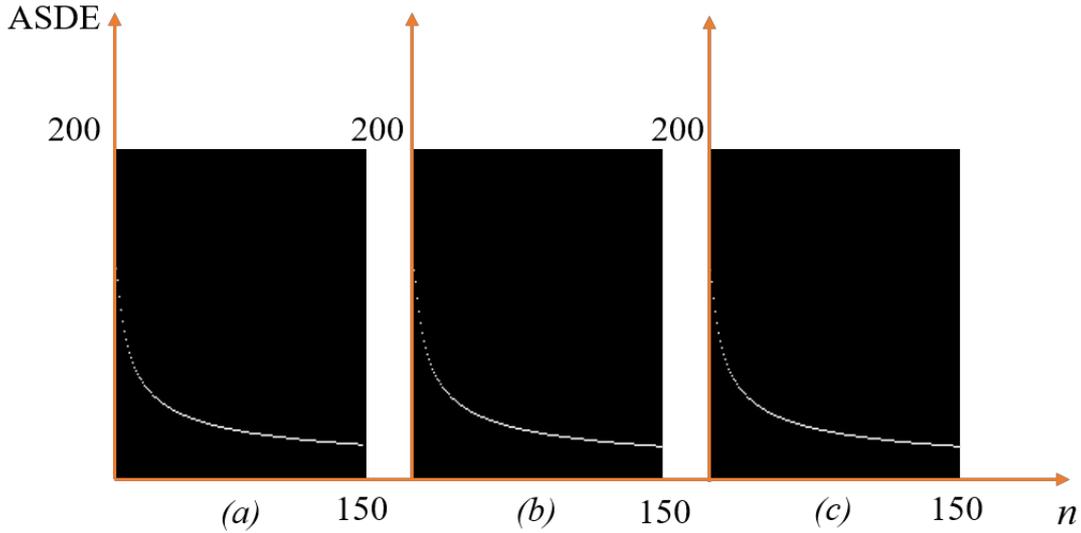

***Figure 4.*** *Plots of ASDE against size of spatiotemporal buffer. (a) is a buffer of resolution 640x480, (b) is of resolution 1280x720, (c) is the plot of proposed function (9). As can be observed, there are hardly any differences between the curves, showing that (1) the curve is unaffected by resolution; (2) our proposed function optimally fits with the relation between buffer size and threshold value.*

We hereby propose a function of threshold value with respect to the size of a buffer:

$$\varphi(n) = \alpha \sqrt{\frac{(maxval - \frac{maxval}{n})^2 + (n-1)(\frac{maxval}{n})^2}{n}}, \quad (9)$$

where $n$ is the size of buffer, $\alpha$ is a variable parameter with an optimal range of [0.15, 0.25], and *maxval* is the maximal value of energy map. Fundamentally, the formula simulates a worse-case scenario where the most radical energy change (from lowest to highest) happens after n consecutive frames. Then we take a fraction (according to parameter $\alpha$) of its corresponding standard deviation, and use the result to bound the size of our spatiotemporal buffer.

With the spatiotemporal buffer established, the uniform energy map is defined as the average of the collection of energy maps. Same seams are then calculated and removed in sequence from all color frames in the container. The employment of spatiotemporal buffer not only speeds up the seam carving process, but also provides more consistency across frames. Slight alternation occurs less frequently, making the video less jittery and more visually appealing.

### 3.3 Improved Visual Saliency Map

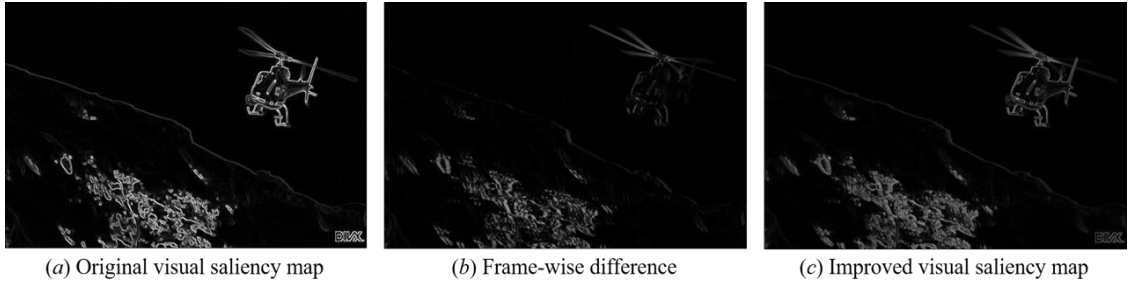

(*a*) Original visual saliency map    (*b*) Frame-wise difference    (*c*) Improved visual saliency map

*Figure 5. Improved visual saliency map.*

Visually, seam removal in video sequences is exceptionally noticeable. The problem is exacerbated with the introduction of spatiotemporal buffer, since motions are more likely to be undercompensated. To protect the consistency of motions, an improved visual saliency map that compensates for frame-wise difference is proposed. Specifically, the energy function takes the absolute difference between the current frame and the previous frame, and blends the result in a weighted manner with the original energy map based on Sobel gradient energy. The weights may vary, but considering the higher priority of motion, more weight should be assigned to difference and less to energy. In our implementation, the weights are 0.6 and 0.4 respectively.

## 4 Experiment Results

All the experiments are conducted on a PC with Intel Core i7 processor running at 3.4 GHz with 16 Gigabytes of RAM. The methods are implemented with Python 2.7 and OpenCV 3.1.0.

For Seam Carving with Parental Labeling, we conducted two experiments to compare its speed performance with the original seam carving method. First, we ran both seam carving methods recursively, reducing the width of previous results by a factor of 0.8 in each recursion. The results reflect the methods' performances under different resolutions. Second, we ran a direct seam carving, reducing the width of images to various target values in one step. The results reflect the methods' performances when removing different number of seams.

For Fast Video Retargeting, we compared our proposition with raw video retargeting which applies original seam carving to separate frames. We first recorded the time consumption of directly retargeting videos to various widths. Then, instead of recursively retargeting, we retargeted the same video of different resolutions to the same extents. Results are shown as follows.

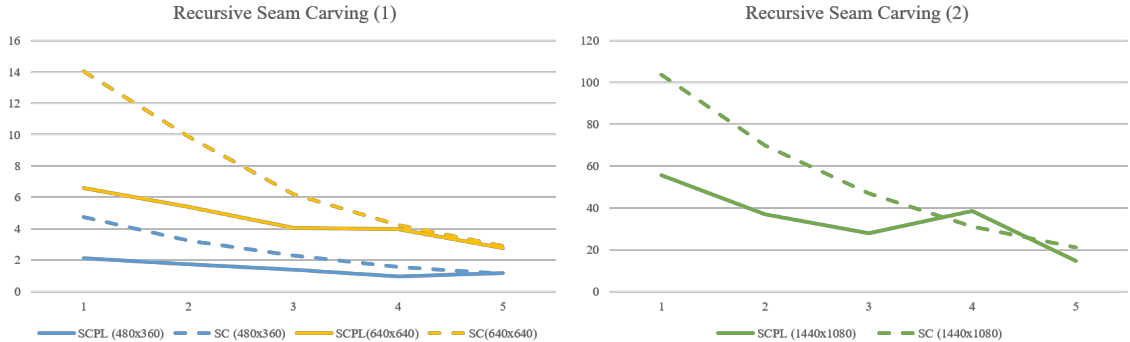

*Figure 6.* Results of recursive seam carving. The horizontal axis is the number of recursive loops. In each loop the image is resized to 0.8 of its previous width. The vertical axis is the performance of each trial measured in seconds.

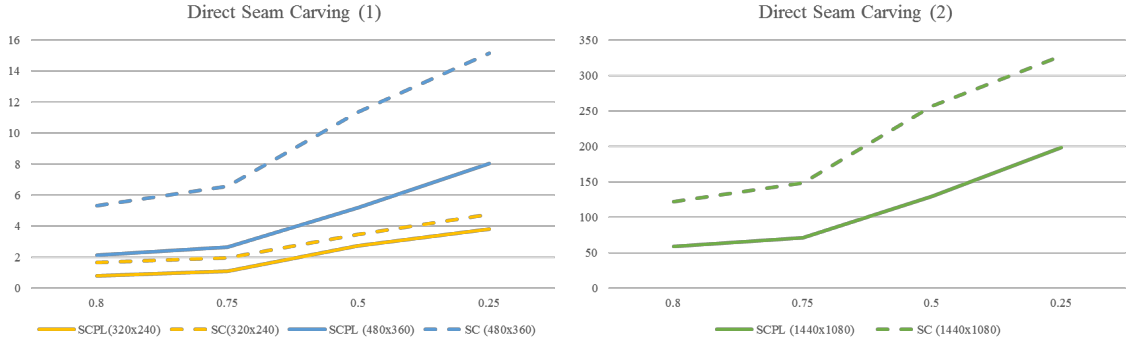

*Figure 7. Results of direct seam carving. The horizontal axis is the ratio of target width to original width. The vertical axis is the performance of each trial measured in seconds.*

Through an interpretation of data, we discover a universal reduction of processing time using SCPL. The reduction is especially significant when resizing images with high resolution, with SCPL running 50% faster. When the resolution is low, however, the run time of SC approaches SCPL. The reason lies in the fact that in our experiments only the width is resized recursively. When the width-to-height ratio becomes low, there are fewer parents per iteration since more seams derive from the same parents.

*Table 1. Performance of Video Retargeting on Various Samples*

| Name | Resolution | Number of frame | Target resolution | Method | Time (second) |
|---|---|---|---|---|---|
| "Big Belly" | 640x480 | 377 | 512x480 | Proposed | 124.36 |
| | | | | Raw | 1255.19 |
| | | | 384x480 | Proposed | 214.24 |
| | | | | Raw | 2484.91 |
| "Lion" | 640x480 | 117 | 512x480 | Proposed | 80.72 |
| | | | | Raw | 600.90 |
| | | | 384x480 | Proposed | 140.88 |
| | | | | Raw | 1225.60 |
| "Lego" | 560x320 | 164 | 448x320 | Proposed | 32.28 |
| | | | | Raw | 263.89 |
| | | | 336x320 | Proposed | 56.21 |
| | | | | Raw | 532.16 |
| "Helicopter" | 720x576 | 178 | 576x576 | Proposed | 143.31 |
| | | | | Raw | 1594.19 |
| | | | 432x576 | Proposed | 282.70 |
| | | | | Raw | 3096.67 |
| | 320x240 | | 256x240 | Proposed | 16.32 |

|  |  |  |  | Raw | 201.71 |
|---|---|---|---|---|---|
|  |  |  | 192x240 | Proposed | 31.44 |
|  |  |  |  | Raw | 426.67 |
|  | 176x144 |  | 140x144 | Proposed | 3.43 |
|  |  |  |  | Raw | 52.65 |
|  |  |  | 105x144 | Proposed | 6.73 |
|  |  |  |  | Raw | 108.34 |

For video retargeting, our proposed method yields a 90.6% reduction of run time on average. The enhancements are more significant for low resolution videos because the majority of time is consumed by operations on the spatiotemporal buffer, whose time complexity is directly related to the frame resolution. When retargeting sample "Helicopter" from 176x144 to 105x144, for instance, our approach achieves a whopping 93.7% reduction of time. It can thus be concluded that our approach has a universally superior performance over raw image retargeting.

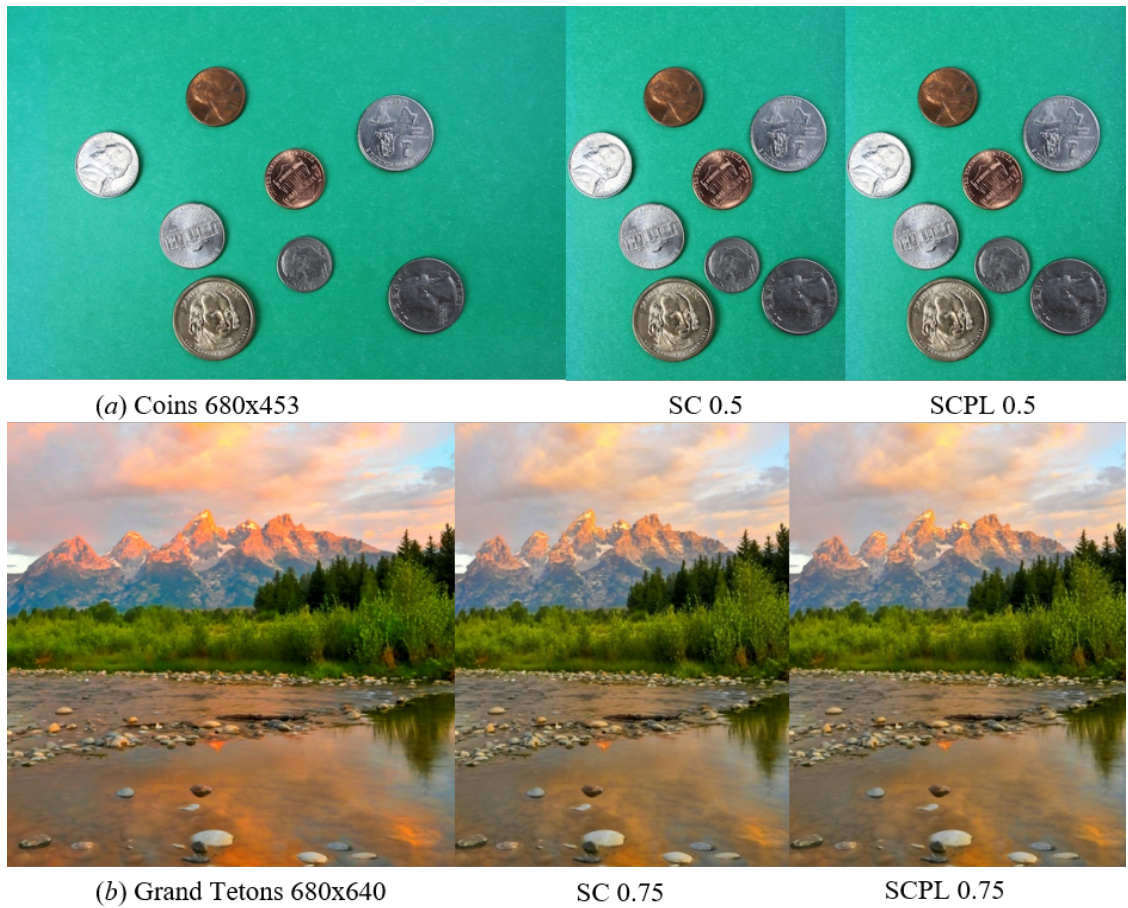

(a) Coins 680x453     SC 0.5     SCPL 0.5

(b) Grand Tetons 680x640     SC 0.75     SCPL 0.75

*Figure 8. Comparison of image quality between SCPL and SC.*

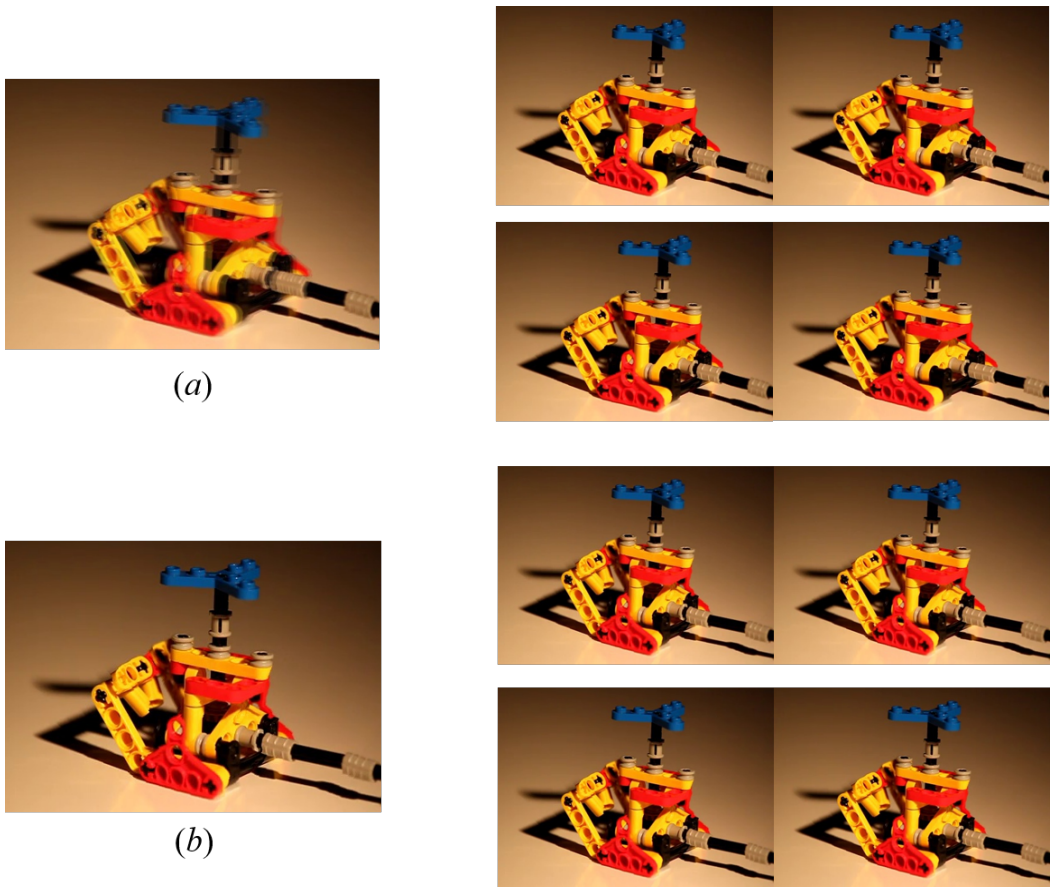

***Figure 9.*** *Blending 4 consecutive frames of both results. (a) is from raw video retargeting, (b) is from Fast Video Retargeting based on SCPL. The original video sequence is nearly static. The result of raw video retargeting is blurry, indicating a displacement of content. Our result yields a sharper image.*

Despite the time reduction, it can be observed that SCPL yields results with qualities similar to if not better than those of original seam carving. Fast video retargeting also achieves similar visual quality to that of raw seam carving. Moreover, due to the spatiotemporal buffers, retargeted videos are significantly more consistent and visually appealing.

## 5  Conclusion

Fast and high-quality video retargeting methods have always been an intriguing research topic. In this paper, we propose a novel approach that has three major contributions to this field of research. Firstly, we propose an improved single-frame seam carving method, SCPL, that labels each seam with its parent and removes the least significant seam from each parent per iteration. Secondly, we propose a spatiotemporal buffer that gathers a period of frames with low energy variation and applies seam carving on the spatiotemporal cube. We present an operator, average standard deviation of energy (ASDE), to determine the size of buffer. Thirdly, we enhance the energy function specifically for video retargeting by compensating frame-wise motions whose destruction causes conspicuous discontinuities and affects visual perception.

It is demonstrated that our proposition greatly outperforms other seam carving and video retargeting methods. However, due to its low parallelism, its performance is largely confined by the resolution of source and the number of seam removals. Also, although jittering is alleviated in the retargeted video, objects may be slightly distorted as the visual saliency map is less representative of a single frame. In the future, it would be promising to further improve performance by avoiding repetitive process through partial energy update and seam memorization. Distortions can be optimized by the employment of better visual saliency maps.